\documentclass[runningheads]{llncs}

\usepackage{eccv}

\usepackage{eccvabbrv}

\usepackage{graphicx}
\usepackage{booktabs}
\usepackage{booktabs, xcolor, colortbl, graphicx}
\usepackage{tcolorbox}
\usepackage{makecell}    % \makecell 多行表头
\usepackage{multirow}    % \multirow（如果用到的话）
\usepackage{xcolor}      % \rowcolor
\usepackage{colortbl}    % 表格颜色支持
\definecolor{best}{rgb}{0.8, 0.0, 0.0}
\definecolor{second}{rgb}{0.0, 0.7, 0.9}
\definecolor{D}{HTML}{FFF9D6}
\newcommand{\first}[1]{\textcolor{best}{\textbf{#1}}}
\newcommand{\secon}[1]{\textcolor{second}{\textbf{#1}}}

\usepackage[accsupp]{axessibility}  % Improves PDF readability for those with disabilities.

\usepackage[pagebackref=true,breaklinks=true,letterpaper=true,colorlinks,bookmarks=false]{hyperref}

\usepackage{orcidlink}

\begin{document}

\title{QualiTeacher: Quality-Conditioned Pseudo-Labeling for \\ Real-World Image Restoration} 

\titlerunning{Abbreviated paper title}

\author{Fengyang Xiao$^{1,*}$\,,
        Jingjia Feng$^{1,}$\thanks{Equal Contribution, $\dagger$ Corresponding Author}\,~,
        Peng Hu$^{2,}$ \,, 
        Dingming Zhang$^{1,}$ \,, \\
        Lei Xu$^{3}$ \,,
	Guanyi Qin$^{4}$\,,
	Lu Li$^5$\,,
	Chunming He$^{1,\dagger}$\,,
        and Sina Farsiu$^6$\\
 }

\authorrunning{Xiao et al.}

\institute{Duke University \and
Tsinghua University \and \'Ecole Polytechnique F\'ed\'erale de Lausanne (EPFL) \and National University of Singapore \and
Sun Yat\text{-}sen University\\
\email{fengyang.xiao@duke.edu} or \email{chunming.he@duke.edu}
}

\maketitle
% \begin{center}
% \includegraphics[width=\textwidth]{Figure/Motivation.pdf}\vspace{-3mm} 
% \captionof{figure}{Comparison of pseudo-label utilization strategies in Mean Teacher framework. 
% % (a) Unconditional trust forces the student to replicate rather than remove teacher artifacts. (b) Aggressive filtering discards most PLs, causing over-smoothed outputs due to limited data diversity. (c) QualiTeacher retains all PLs with IQA-based quality scores as conditions, enabling the student to learn a quality-graded restoration manifold. 
% } \label{Fig:ArchitectureCompare}
% \end{center}
% \vspace{-10pt}
\begin{abstract}
Real-world image restoration (RWIR) is a highly challenging task due to the absence of clean ground-truth images. Many recent methods resort to pseudo-label (PL) supervision, often within a Mean-Teacher (MT) framework, where a teacher network generates targets for a student network. However, these methods face a critical paradox: unconditionally trusting the often imperfect, low-quality PLs forces the student model to learn undesirable artifacts, while discarding them severely limits data diversity and impairs model generalization. In this paper, we propose QualiTeacher, a novel framework that transforms pseudo-label quality from a noisy liability into a conditional supervisory signal. Instead of filtering, QualiTeacher explicitly conditions the student model on the quality of the PLs, estimated by an ensemble of complementary non-reference image quality assessment (NR-IQA) models spanning low-level distortion and semantic-level assessment. This strategy teaches the student network to learn a quality-graded restoration manifold, enabling it to understand what constitutes different quality levels. 
Consequently, it can not only avoid mimicking artifacts from low-quality labels but also extrapolate to generate results of higher quality than the teacher itself. 
To ensure the robustness and accuracy of this quality-driven learning, we further enhance the process with a multi-augmentation scheme to diversify the PL quality spectrum, a score-based preference optimization strategy inspired by Direct Preference Optimization (DPO) to enforce a monotonically ordered quality separation, and a cropped consistency loss to prevent adversarial over-optimization (reward hacking) of the IQA models. Experiments on standard RWIR benchmarks demonstrate that QualiTeacher can serve as a plug-and-play strategy to improve the quality of the existing pseudo-labeling framework, establishing a new paradigm for learning from imperfect supervision. Code will be released.
% We further enhance this process with a multi-augmentation scheme to diversify the PL quality spectrum, a cropped consistency loss to avoid overoptimization, and a score-based preference optimization strategy inspired by DPO to lead the student model to separate different output quality spaces. Experiments on standard RWIR benchmarks demonstrate that QualiTeacher can serve as a plug-and-play strategy to improve the quality of the existing pseudo-labeling framework, establishing a new paradigm for learning from imperfect supervision. Code will be released.
\end{abstract}

%% narrow the gap between equations and sentences
\setlength{\abovedisplayskip}{2pt}
\setlength{\belowdisplayskip}{2pt}
\section{Introduction}
\label{introduction} %\vspace{-2mm}
Real-world degradations are complex and spatially varying, encompassing an often unpredictable combination of environmental noise, blur, compression artifacts, and sensor noise~\cite{he2023degradation,fang2024real,chen2023hierarchical}. The primary challenge in tackling this real-world image restoration (RWIR) problem lies in the inability to obtain clean, high-quality ground-truth images corresponding to the degraded inputs.

\begin{figure}[t]
\centering
\setlength{\abovecaptionskip}{0cm} 	\setlength{\belowcaptionskip}{0cm}
\includegraphics[width=\textwidth]{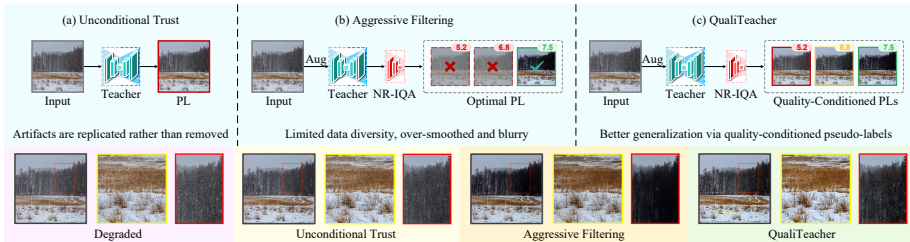}
 \vspace{-3mm} 
\caption{Comparison of pseudo-label (PL) utilization strategies in the mean-teacher framework. \textbf{(a) Unconditional Trust:} PLs are used without filtering, causing degradation artifacts to be replicated by the student. \textbf{(b) Aggressive Filtering:} Low-quality PLs are discarded via NR-IQA filtering, yet over-smoothed outputs that receive deceptively high scores survive, introducing blurriness. \textbf{(c) QualiTeacher (Ours):} Nearly all PLs are retained (with only extreme outliers discarded), with their quality scores injected as continuous conditioning signals, enabling the student to leverage full data diversity while remaining aware of each sample's reliability.}

\label{Fig:ArchitectureCompare}\vspace{-6mm}
\end{figure}

To circumvent this, a popular and effective paradigm is the mean-teacher (MT) framework, which leverages self-supervised or semi-supervised learning. In this setup, a teacher model (often an exponential moving average of the student) generates pseudo-labels (PLs) to supervise the student model. Recent works have focused on refining this process, for instance, by creating ``label pools'' to store the best historical PLs, with the implicit goal of filtering supervision to include only high-quality targets~\cite{fang2024real}.

Despite their success, they suffer from a fundamental dilemma regarding the quality of the PLs. Owing to the limited capacity of the teacher model, these PLs unavoidably contain residual artifacts or blurred textures. Existing methods thus face a critical paradox, shown in \cref{Fig:ArchitectureCompare}.
% \textbf{(1) Unconditional Trust:} Forcing the student to naively regress to the imperfect PLs causes it to learn the teacher's mistakes, leading to a sub-optimal domain shift where artifacts are replicated rather than removed.
% \textbf{(2) Aggressive Filtering:} 
% % Low-quality pseudo-labels are discarded via NR-IQA thresholding, which not only reduces data diversity, but also introduces selection bias—since over-smoothed outputs tend to receive favorable IQA scores and thus dominate the retained set, further pushing restorations toward blurriness.
%  Discarding all but the highest-quality PLs starves the model of training data, reducing data diversity and inhibiting its ability to generalize to unseen, complex degradations.
This ``all-or-nothing'' approach severely limits the student's potential, forcing it to choose between learning from corrupted data or learning from scarce data. In this work, we argue that this dilemma stems from a flawed premise. Instead of treating low-quality PLs solely as noise to be filtered, we should treat them as valuable, graded learning signals to be learned from.

Motivated by this insight, we propose QualiTeacher, a novel framework that transforms this challenge into an opportunity. Our core idea is to make the student model quality-aware. Instead of just learning what to restore, the student learns how the quality of the restoration relates to the underlying content. Specifically, we employ off-the-shelf non-reference Image Quality Assessment (NR-IQA) models to score each PL. This quality score is not used for filtering; rather, it is embedded (using a continuous embedding, \textit{e.g.}, an MLP or sinusoidal encoding) and injected into the student network as a quality condition.

This quality-conditioned learning strategy achieves two critical goals. First, it allows the student to intelligently discount artifacts from low-quality PLs (\textit{e.g.}, ``I am told this target is a 6/10 image, so I will not fully trust its noisy texture''). Second, by training on a spectrum of qualities, the student learns the entire quality-graded restoration manifold. This enables it to extrapolate beyond the teacher's best performance. Even if the teacher only produces 5-6/10 quality labels, the student learns the vector direction of ``quality improvement'' and can generate restorations at a quality level exceeding the teacher's typical output range at inference time by being fed a high-quality score condition.
% can generate 9/10 quality results at inference time by being fed a high-quality score condition. 
This ability to actively generate restorations corresponding to specific target scores is the key to breaking the upper bound of the teacher model.

However, merely injecting scores as conditions does not guarantee that the student explicitly learns a monotonic mapping between the condition space and the output quality. To facilitate this, we first propose a multi-augmentation strategy for the teacher. Its goal is not to find the single best label, but to create a rich and diverse curriculum of PLs across a wide quality spectrum, providing the student with varied training data. To address non-uniform quality within an image, we introduce a spatial-aware IQA weighting mechanism to guide the student to focus on high-quality regions. Crucially, to ensure the conditional student produces well-separated outputs proportional to the injected score, we introduce a score-based preference optimization objective inspired by Direct Preference Optimization~\cite{rafailov2023direct}. By treating NR-IQA scores as automated preference signals, we bypass the need for human annotation and directly optimize the deterministic student model to establish a discriminative, ordered quality manifold. Furthermore, unconstrained optimization toward higher IQA scores may lead to adversarial over-optimization, where the model exploits local statistical shortcuts to inflate scores without genuine perceptual improvement (a phenomenon analogous to reward hacking). To mitigate this, we design a cropped consistency loss that enforces spatial uniformity of quality improvement, thereby ensuring the reliability and robustness of the learned score-quality mapping.

Our main contributions are summarized as follows:

\noindent (1) We propose QualiTeacher for the RWIR task, a novel framework that leverages pseudo-label quality as a condition for deep quality understanding.
% and extrapolation.

 \noindent (2) We propose a score-based preference optimization strategy inspired by Direct Preference Optimization to explicitly learn discriminative mappings between score conditions and output quality spaces.
% To our knowledge, it is the first to leverage preference optimization in non-generative low-level vision.}
% tasks. 
% demonstrate that this quality-conditioned approach enables the student model to extrapolate, producing restoration results superior to the teacher's maximum quality.

\noindent (3) Experiments verify that QualiTeacher serves as a plug-and-play strategy to improve existing pseudo-labeling frameworks, achieving new SOTAs on several RWIR benchmarks across diverse degradation types.

\section{Related Works}
\vspace{-1mm}
\subsection{Semi-Supervised Image Restoration}\vspace{-1mm}
RWIR lacks clean ground-truth for real degraded inputs, motivating semi-supervised paradigms. Existing approaches leverage unpaired data via cycle-consistency~\cite{wu2023semi, wu2023mixcycle} or contrastive learning~\cite{basak2023pseudo,yang2022class}, while more recent methods adopt pseudo-label-based strategies~\cite{wang2022semi} where a pre-trained model generates supervision targets. Despite their effectiveness, these methods treat all pseudo-labels uniformly or apply binary filtering, without exploiting the continuous quality spectrum as a learning signal. QualiTeacher fundamentally differs by re-purposing pseudo-label quality as a conditional input rather than a filtering criterion.
\vspace{-1mm}
\subsection{Mean-Teacher Frameworks}\vspace{-1mm}
The mean-teacher (MT) framework~\cite{tarvainen2017mean} maintains an EMA of the student as the teacher to produce stable pseudo-labels, and has been adapted to desnowing~\cite{lai2025snowmaster}, dehazing~\cite{fang2024real}, and other general restoration tasks~\cite{xu2024towards, long2025semiddm}. Recent works further improve the pipeline through data augmentation~\cite{fang2024real}, consistency regularization~\cite{xu2023ambiguity}, and adaptive label selection~\cite{liu2023mixteacher}. However, existing MT-based methods focus on improving pseudo-label reliability through better filtering, while overlooking the quality information embedded in each pseudo-label. QualiTeacher injects the quality score directly into the student, transforming teacher-student learning from unconditional mimicry to quality-conditioned generation.
\vspace{-1mm}
\subsection{NR-IQA in Low-Level Vision}\vspace{-1mm}
No-reference image quality assessment (NR-IQA) estimates perceptual quality without a reference image, ranging from statistics-based methods like BRISQUE~\cite{mittal2012no} to learning-based approaches such as MUSIQ~\cite{Ke_2021_ICCV} and CLIP-IQA~\cite{wang2023exploring}. In low-level vision, NR-IQA scores have primarily served as post-hoc evaluation metrics~\cite{he2023reti}, with limited use as auxiliary losses~\cite{zhang2025augmenting} or for dataset curation~\cite{dai2023emu}. None of these methods exploits NR-IQA scores as an explicit condition injected into the restoration network. Our work is the first to systematically integrate NR-IQA into every stage of the pipeline, from pseudo-label curation to score-conditioned generation to score-based preference optimization.

\vspace{-1mm}
\subsection{Preference Optimization}\vspace{-1mm}
Direct Preference Optimization (DPO)~\cite{rafailov2023direct} bypasses explicit reward modeling by directly optimizing a policy from preference pairs, and has been widely adopted for large language models~\cite{xiao2024cal}. In the visual domain, DPO has been applied to diffusion models for text-to-image generation~\cite{Wang_2025_CVPR} and video synthesis~\cite{liu2025videodpo}. However, preference optimization remains unexplored in the non-generative low-level image restoration tasks. We adapt a score-based preference optimization that bypasses the theoretical obstacle of deterministic models lacking probability outputs, using NR-IQA scores as preference signals and naturally constructing pairs through the score-conditioned strategy.

\section{Methodology}

\subsection{QualiTeacher}

\begin{figure}[t!]
    \centering
    \setlength{\abovecaptionskip}{0cm} 	\setlength{\belowcaptionskip}{0cm}
    \includegraphics[width=1.0\linewidth]{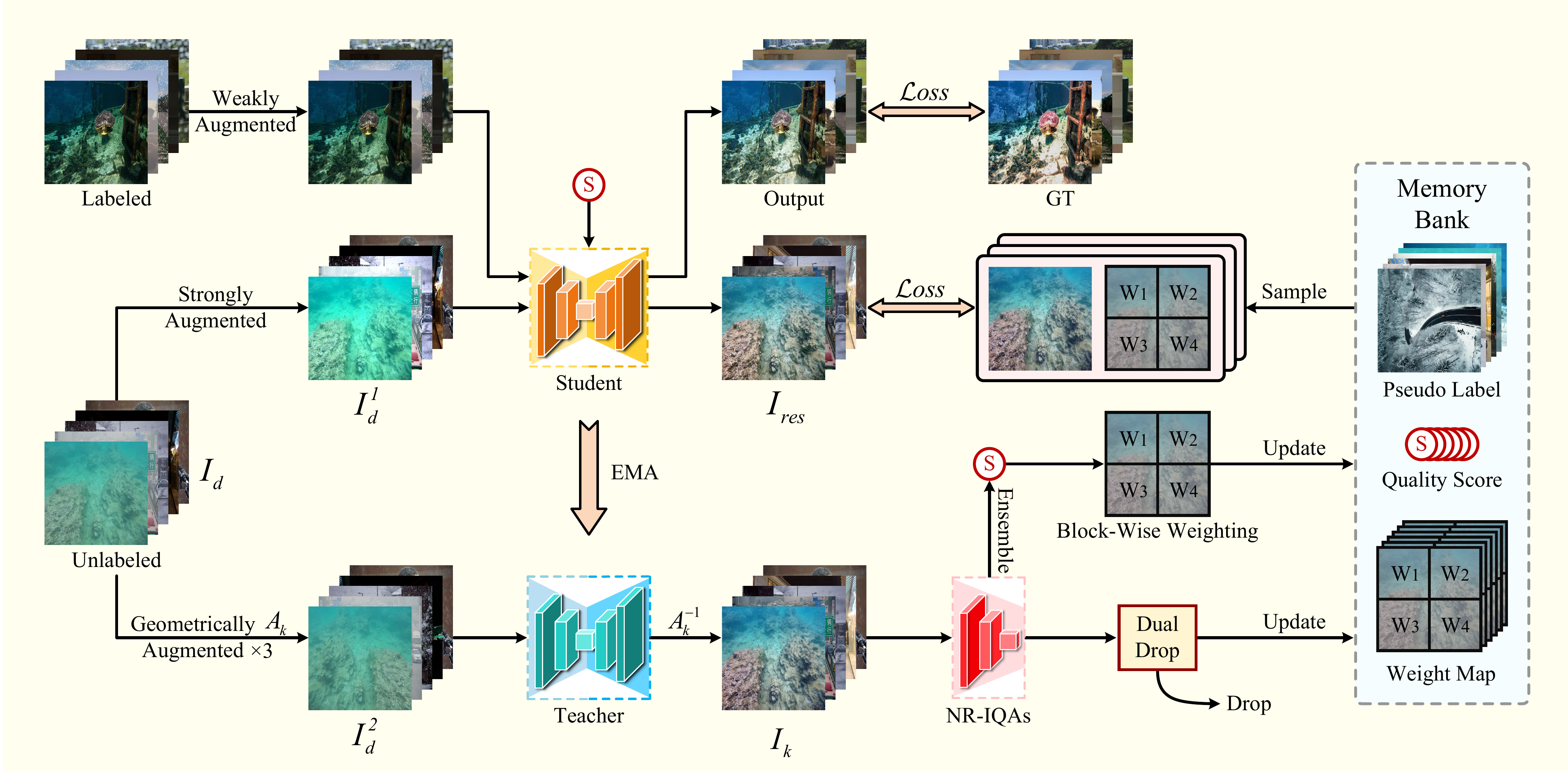}
    \caption{Overall framework of QualiTeacher. 
    By re-purposing pseudo-label quality as a conditional signal rather than a filtering criterion, the student network can learn a quality-graded manifold and extrapolate beyond the teacher's capabilities.
    }
    \label{fig:placeholder}
    \vspace{-15pt}
\end{figure}

The complete pipeline of QualiTeacher is illustrated in \cref{fig:placeholder}. The framework integrates three synergistic components: (i)~a \textit{quality-based pseudo-labeling} module (Sec.~\ref{sec.3.2}) that curates a diverse spectrum of quality-graded PLs via multi-augmentation, multi-IQA consensus scoring, and dual-drop gating; (ii)~a \textit{quality-aware score injection and weighting} module (Sec.~\ref{sec.3.3}) that embeds IQA scores into the student network as explicit quality conditions with spatial-aware weighting; and (iii)~a \textit{quality-driven optimization} module (Sec.~\ref{sec.3.4}) that incorporates score-based preference optimization and cropped consistency regularization to ensure faithful quality controllability. We elaborate on each component below.

% QualiTeacher builds upon the mean-teacher (MT) architecture but takes a fundamentally different stance toward pseudo-label quality: rather than unconditionally trusting imperfect PLs or aggressively discarding low-quality ones, it treats PL quality as a condition to be learned from. The complete pipeline is illustrated in Fig.X.

% Specifically, the framework integrates three synergistic components: (i)~a \textit{quality-based pseudo-labeling} module (Sec.~\ref{sec.3.2}) that employs multi-augmentation, multi-IQA consensus scoring, and a dual-drop gating mechanism to curate a diverse spectrum of quality-graded PLs; (ii)~a \textit{quality-aware score injection and weighting} module (Sec.~\ref{sec.3.3}) that embeds the IQA score into the student network as an explicit quality condition with spatial-aware weighting; and (iii)~a \textit{quality-driven optimization} module (Sec.~\ref{sec.3.4}) that incorporates score-based preference optimization and cropped consistency regularization to ensure faithful quality controllability. Together, these components enable the student to learn a quality-graded restoration manifold, allowing it to not only discount artifacts from low-quality PLs but also generate results conditioned on higher target scores at inference time. In the following, we elaborate on each component.

\subsection{Quality Based Pseudo-Labeling} \label{sec.3.2}

\noindent \textbf{Augmentation and Scoring.} 
In the standard MT paradigm, the teacher generates a single pseudo-label per input, yielding a narrow quality distribution that limits supervision diversity. To broaden the quality spectrum of PLs and enhance framework robustness, we introduce a multi-augmentation strategy leveraging geometric transformations to elicit diverse restoration outputs from the teacher. Specifically, given a degraded input $I_d$, we apply three complementary geometric augmentations (horizontal flip, vertical flip, and $90^{\circ}$ rotation), to produce augmented variants $\{\mathcal{A}_k(I_d)\}_{k=1}^{3}$. Each variant is independently processed by the teacher $f_{\theta_T}$, and the outputs are mapped back to the original coordinate space via corresponding inverse transformations, yielding three PLs:
\begin{equation}\label{eq:aug}
    I_k = \mathcal{A}_k^{-1}\!\Big(f_{\theta_T}\!\big(\mathcal{A}_k(I_d)\big)\Big), \quad k \in \{1, 2, 3\},
\end{equation}
where $\mathcal{A}_k^{-1}$ denotes the inverse of the $k$-th geometric transformation.

Although these augmentations are theoretically lossless, the teacher responds differently to each spatial configuration, naturally producing PLs of varying quality. To evaluate perceptual quality, we employ complementary NR-IQA models across levels: low-level metrics (MUSIQ-KonIQ~\cite{Ke_2021_ICCV}, BRISQUE~\cite{mittal2012no}) capture perceptual distortions, while high-level metrics (CLIP-IQA~\cite{wang2023exploring}) leverage vision-language priors for semantic-aware quality estimation. This multi-IQA strategy ensures that quality scores reflect comprehensive and robust assessment rather than the bias of any single metric.

% \noindent \textbf{Dual-Drop Mechanism}\quad
% To ensure both multi-IQA consistency and augmentation robustness, we introduce a dual-drop mechanism. In the first drop stage, multiple IQA models jointly evaluate a single PL, and the variance across their normalized scores is computed. Formally, given $M$ IQA models, a PL $\hat{I}_k$ is considered consistent if:
% \begin{equation}\label{eq:drop1}
%     \mathrm{Var}\!\Big(\big\{\bar{S}_m(\hat{I}_k)\big\}_{m=1}^{M}\Big) < \tau_1, \quad \forall\, k \in \{1, 2, 3\},
% \end{equation}
% where $\bar{S}_m$ denotes the min-max normalized score of the $m$-th IQA model and $\tau_1$ is the consistency threshold. If any PL fails to satisfy this criterion, the entire set of augmented PLs is dropped, thereby enforcing multi-IQA consistency and ensuring perceptual quality coherence from both low-level and high-level perspectives. In the second drop stage, each IQA model independently evaluates the entire set of PLs, verifying that the augmentation process yields robust and stable quality estimates:
% % \begin{equation}\label{eq:drop2}
% %     \max_k \bar{S}_m(\hat{I}_k) - \min_k \bar{S}_m(\hat{I}_k) < \tau_2, \quad \forall\, m \in \{1, \ldots, M\},
% % \end{equation}
% \begin{equation}
%      \mathrm{Var}\!\Big(\big\{\bar{S}_m(\hat{I}_k)\big\}_{m=1}^{M}\Big) < \tau_2, \quad \forall\, k \in \{1, 2, 3\},
% \end{equation}
% where $\tau_2$ controls the threshold for score variation across different geometric transformations.
\noindent \textbf{Dual-Drop Mechanism.}
To ensure multi-IQA consistency and augmentation robustness, we introduce a dual-drop mechanism. In the first drop stage, multiple IQA models jointly evaluate each PL, and the variance across their normalized scores is computed. Given IQA models, $\forall\, k, m \in \{1, 2, 3\}$, $I_k$ is consistent if:
\begin{equation}\label{eq:drop1}
    \mathrm{Var}\!\Big(\big\{\bar{S}_m(I_k)\big\}_{m=1}^{3}\Big) < \tau_1,  \quad
     \mathrm{Var}\!\Big(\big\{\bar{S}_m(I_k)\big\}_{k=1}^{3}\Big) < \tau_2, 
     % \quad \forall\, k, m \in \{1, 2, 3\}.
\end{equation}
where $\bar{S}_m$ denotes the min-max normalized score of the $m$-th IQA model and $\tau_1$ is the consistency threshold. If any PL fails this criterion, the entire augmented set is dropped, enforcing perceptual quality coherence from both low-level and high-level perspectives. Then, each IQA model independently evaluates the full set of PLs, verifying that augmentation yields stable quality estimates, and $\tau_2$ controls the robustness for score variation across geometric transformations.

\noindent \textbf{Gating Mechanism.}
To ensure high-quality supervision, we maintain a memory bank $\mathcal{B}$ defined to store the PLs. 
The bank is dynamically updated and retains only the top-3 pseudo-labels from the training history. During training, not all PLs are admitted into the memory bank; only those that pass the dual-drop criterion in each round serve as candidates $\mathcal{C}$ for the gating mechanism. Specifically, the existing PLs in the bank and the candidate PLs are collectively retrieved and re-ranked, from which the new top-3 are selected:
\begin{equation}\label{eq:gate}
    \mathcal{B} \leftarrow \mathrm{Top\text{-}3} \big(\mathcal{B} \cup \mathcal{C},\; S\big),
\end{equation}
where the ranking is determined by the ensemble score $S$ of MUSIQ-KonIQ~\cite{Ke_2021_ICCV}, BRISQUE~\cite{mittal2012no}, and CLIP-IQA~\cite{wang2023exploring}, to :
 \begin{equation}\label{eq1}
     S= 0.4 S_{MUSIQ-KonIQ}+0.4 S^{'}_{BRISQUE}+ 0.2 S_{CLIP-IQA},
 \end{equation}
where $S^{'}_{BRISQUE}=(100-{S}_{BRISQUE})$ is the inverted BRISQUE score (converting lower-is-better to higher-is-better), after which all NR-IQA scores are normalized to [0, 1].

% Since BRISQUE is a lower-is-better metric with a range of [0, 100], we convert it to $\bar{S}_{BRISQUE}=(100-{S}_{BRISQUE})$ to align with the other metrics, and then normalize all NR-IQA scores to [0, 1].

\subsection{{Quality-Aware Score Injection and Weighting}} \label{sec.3.3}
\noindent \textbf{Quality-Aware Score Injection.}
A central design principle of QualiTeacher is to condition the student network on the perceptual quality of each pseudo-label, transforming restoration from an unconditional mapping into a quality-aware generation process.
To achieve this, we inject the aggregated IQA score into the student network at the feature level via a simple yet effective additive mechanism. The conditional student model learns to capture the relationship between PLs and their associated quality scores. During training, when the highest quality score is used as the condition, the PL is set to the highest quality image, resembling class-conditional generation and enabling the model to learn quality-specific capabilities across diverse levels. {Concretely, given the ensemble quality score $S \in \mathbb{R}$ of a pseudo-label, we first map it into a high-dimensional embedding $\mathbf{e}_S \in \mathbb{R}^{C}$ through a score embedding module:}
\begin{equation}\label{eq:embed}
    \mathbf{e}_S = \mathrm{MLP}\!\big(\gamma(S)\big),
\end{equation}
\vspace{-7pt}
\begin{equation}
    \gamma(S) = \big[\sin(2^0\pi S),\, \cos(2^0\pi S),\, \ldots,\, \sin(2^{L-1}\pi S),\, \cos(2^{L-1}\pi S)\big].
\end{equation}
where $C$ matches the channel dimension of target feature maps and $L$ is the number of frequency bands. The embedding $\mathbf{e}_S$ is then added to intermediate feature representations at selected injection points:
\begin{equation}\label{eq:inject}
    \mathbf{F}' = \mathbf{F} + \mathbf{e}_S,
\end{equation}
where $\mathbf{F}$ denotes the feature at the injection layer, and $\mathbf{e}_S$ is broadcast along spatial dimensions. For instance, in U-Net, injection points are chosen at the encoder-decoder bottleneck, where feature maps carry the most compact and semantically rich representations.
More generally, this additive strategy is architecture-friendly with negligible computational overhead, ensuring QualiTeacher serves as a plug-and-play module for various restoration backbones.

% Taking CoRUN~\cite{} as an example, which adopts a 3-layer U-Net, we inject the score embedding at the second encoder layer and the zeroth decoder layer---i.e., the two layers adjacent to the U-Net bottleneck. This design is motivated by two considerations: (1)~injecting at the bottleneck ensures that the quality condition modulates the most abstract feature representations, thereby influencing both the decoding and reconstruction process globally; and (2)~injecting at shallow layers would risk interfering with low-level spatial details that should remain quality-agnostic, while injecting only at the deepest layer may limit the condition's influence on the decoder pathway. By placing the injection at both sides of the bottleneck, the quality signal effectively bridges the encoder and decoder, guiding the student to produce outputs whose perceptual quality aligns with the specified score condition. 

\noindent {\textbf{Block-wise Evaluation and Weighted Masking.}}
Pseudo-label quality is often spatially non-uniform: certain regions may be well-restored while others retain visible artifacts. To account for such intra-image quality variation, we propose a block-wise evaluation and weighted masking strategy. Specifically, we partition each pseudo-label $I_k$ into a regular grid of $2 \times 2$ non-overlapping blocks $\{B_{i,j}\}$, and evaluate the quality of each block independently using the ensemble IQA score~\cref{eq1}.
 The resulting weight map $\mathbf{W} = \{w_{i,j}\}$ is upsampled to the full resolution and applied element-wise to the pixel-level reconstruction loss, guiding the student to focus on regions where the pseudo-label is most reliable while reducing the influence of artifact-prone areas. 

\subsection{\textbf{Quality-Guided Optimization}} \label{sec.3.4}

\noindent \textbf{Score-Based Preference Optimization.}
In the preceding training stages, the student model has learned to perform restoration conditioned on different quality scores. However, it merely fits the teacher's pseudo-labels under varying score conditions, without explicitly learning the monotonic mapping between score spaces and output quality. As a result, the learned score spaces may lack sufficient discriminability, leading to outputs generated under different score conditions that may not exhibit quality differences.

\begin{figure}[t!]
% \vspace{-10pt}
    \centering
    \setlength{\abovecaptionskip}{0cm} 	\setlength{\belowcaptionskip}{0cm}
    \includegraphics[width=1.0\linewidth]{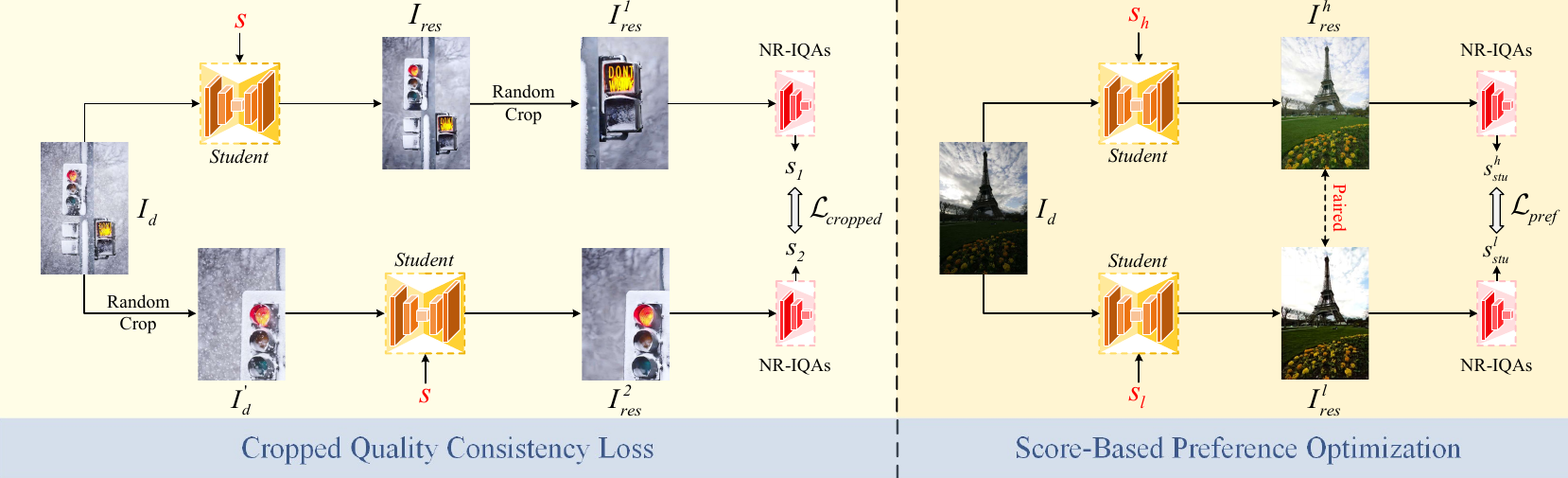}
    % \vspace{-15pt}
    \caption{Overview of the quality-driven optimization strategy in QualiTeacher. Left: Cropped Quality Consistency Loss. Right: Score-Based Preference Optimization.}
    \label{fig:loss}
    \vspace{-0.5cm}
\end{figure}

To address this, we draw inspiration from Direct Preference Optimization (DPO)~\cite{rafailov2023direct}, originally proposed for aligning language models with human preferences. In our setting, NR-IQA scores serve as an automated preference signal that eliminates the need for human annotation: outputs with higher IQA scores are naturally treated as preferred, while those with lower scores are treated as rejected. Moreover, the score-conditioned student can inherently generate outputs of varying quality by injecting different scores, thereby automatically constructing preference pairs, as shown in \cref{fig:loss}

A key theoretical challenge in adapting DPO is that deterministic regression models do not define a probability distribution over outputs, whereas standard DPO requires computing log-probabilities. We bypass this by operating in the output quality space. Specifically, our score-conditioned architecture naturally produces distinct outputs $y^h$ and $y^l$ by injecting different quality scores, constructing preference pairs without stochastic sampling or likelihood estimation. The NR-IQA evaluator then acts as a deterministic reward function mapping each output to a scalar quality measure, replacing log-probabilities in the original DPO formulation. This shifts the optimization target from ``making the preferred output more likely'' to ``making the high-score output measurably better,'' enabling preference optimization in a fully deterministic pipeline.

Unlike standard DPO, which optimizes a single preference direction, our formulation establishes a score-conditioned preference ordering: the objective is not merely to prefer one output over another, but to ensure that restoration quality varies monotonically with the injected score condition, forming a well-separated quality manifold. ~\cref{eq:pref} explicitly forces the student to learn score-space-specific features under different score conditions, producing outputs with clearly distinguishable and ordered quality levels.
\begin{equation}
    \begin{split}
    \mathcal{L}_{pref} = & \underbrace{-\log\sigma\!\Big(\beta\big(S_{stu}(y^h, x) - S_{stu}(y^l, x) - \delta\big)\Big)}_{\mathcal{L}_1} \\
    & - \underbrace{\log\sigma\!\Big(\beta\big(S_{stu}(y^h, x) - S_{tea}(I_k^l, x)\big)\Big)}_{\mathcal{L}_2} 
    + \underbrace{\frac{1}{2} \sum_{n=1}^{N} \|W_n - W_n^{\text{init}}\|_F^2}_{\mathcal{L}_{reg}},
    \end{split}
    \label{eq:pref}
\end{equation}
where $S_{stu}(y, x)$ denotes the IQA-evaluated quality of the student's output $y$ given input $x$, $S_{tea}(I_k^l, x)$ denotes the quality of the teacher's poorest pseudo-label, $y^h$ and $y^l$ are the student's outputs conditioned on high and low quality scores respectively, $\sigma(\cdot)$ is the sigmoid function, $\beta$ is a temperature controlling preference sharpness, and $\delta$ is a margin specifying the minimum desired quality gap between score spaces. $W_n$ is the current weights of the $n$ th score injection convolution layer. $\|\cdot\|_F^2$ denotes the squared Frobenius norm.

\noindent \textbf{Loss 1: $\mathcal{L}_1$}. 
This term enforces inter-space separation: it encourages the student to produce outputs with clearly distinguishable quality when conditioned on different scores, by penalizing cases where the quality gap between $y^h$ and $y^l$ falls below the margin $\delta$. This operates within the student's output space, ensuring that the model learns distinct score spaces and corresponding features.

\noindent \textbf{Loss 2: $\mathcal{L}_2$}. 
While $\mathcal{L}_1$ enlarges the gap between score spaces, it does not specify the \textit{direction} of separation, that is, the model may achieve a large gap by degrading the low-score output rather than improving the high-score one. To prevent this, $\mathcal{L}_2$ introduces a soft lower bound on the student's high-score output, anchoring the separation upward: the student must improve its best output rather than merely worsening its worst. Together, $\mathcal{L}_1$ and $\mathcal{L}_2$ are complementary: $\mathcal{L}_1$ ensures different score conditions produce distinguishable outputs, while $\mathcal{L}_2$ guarantees this separation progressively pushes overall quality upward.

\noindent \textbf{Loss 3: $\mathcal{L}_{reg}$}.
Directly optimizing the preference objective can interfere with underlying restoration capabilities; that is, the image may score higher with the IQA model yet visually contain artifacts. We design $\mathcal{L}_{reg}$ to anchor the injection layer weights near their initial values, ensuring that preference optimization improves quality only through slight, quality-relevant modifications without significantly altering the overall network behavior.

% risks over-optimization---the model may aggressively alter its feature representations to maximize the preference signal, drifting far from the well-established restoration capability and producing outputs with undesirable artifacts. This problem is analogous to the reward hacking phenomenon observed in large-scale RLHF systems. \textcolor{red}{Inspired by L2-SP regularization~\cite{li2020baseline} and RoPO \cite{}, we introduce a weight-space regularization term that anchors the score injection layers to their initial state. Specifically, the $L$ convolutional layers used for score injection are initialized with identity kernels at the beginning of preference optimization, and their initial weights $W_l^{\text{init}}$ are saved as a fixed reference. The regularization term is defined as $\mathcal{L}_{reg}$
% where $\|\cdot\|_F^2$ denotes the squared Frobenius norm. This term constrains the preference optimization to operate within a trust region around the initial model, ensuring that the quality improvement is achieved through gentle, quality-relevant modulations rather than drastic weight shifts that could compromise restoration fidelity. }
\noindent \textbf{Cropped Quality Consistency Loss.}
Using NR-IQA scores as optimization objectives is promising but inherently unstable: without a reference, the model may generate outputs that deceive IQA models into assigning high scores while exhibiting low quality, a phenomenon analogous to adversarial examples and reward hacking in large-scale reward learning systems. To mitigate this, as shown in \cref{fig:loss},  we propose a cropped quality consistency loss that enforces spatial robustness of the quality evaluation. The key idea is to compare two processing orders: (i) restore the full image, then randomly crop a quarter of the restored image and evaluate to obtain $S_1$ via Eq.~(\ref{eq1}); (ii) randomly crop the degraded input first, then restore and evaluate to obtain $S_2$. The loss is defined as:
\begin{equation}
    \mathcal{L}_{cropped} = \|S_1 - S_2\|_1.
    \label{eq:cropped}
\end{equation}

If the model exploits local statistical shortcuts, such as inserting imperceptible artifacts, to inflate global IQA scores, these artifacts are typically spatially unstable and will manifest as discrepancies between $S_1$ and $S_2$. Furthermore, NR-IQA models often exhibit scale-dependent behavior, producing reliable assessments on full images but inconsistent evaluations on local patches. By enforcing consistency between the two processing orders, this loss prevents the model from exploiting scale-dependent biases of NR-IQA models, ensuring that quality improvements are genuine and spatially uniform.

\vspace{-1mm}
\subsection{{Training Strategy}}\vspace{-1mm}

The teacher network parameters $\theta_T$ are updated as an exponential moving average (EMA) of the student parameters $\theta_S$ (with $\alpha=0.998$):
\begin{equation}\label{eq:ema}
    \theta_T \leftarrow \alpha\, \theta_T + (1 - \alpha)\, \theta_S.
\end{equation}
% where $\alpha$ is the EMA decay coefficient and . 

 In the training stage, we train the student model with score injection, score-based preference optimization, and cropped consistency loss, enabling it to acquire quality awareness, restoration ability, and structural fidelity, learning how to map degraded inputs to a different manifold. The objective is:
\begin{equation}\label{eq:stage2}
    \mathcal{L} = \mathcal{L}_{rec} +  \mathcal{L}_{per} + \mathcal{L}_{pref} + \mathcal{L}_{cropped},
\end{equation}
where $\mathcal{L}_{rec} = \mathbf{W}\odot\|f_{\theta_S}(I_d, S) - I_k^*\|_1$ is the pixel-level $\ell_1$ reconstruction loss between the student output and the corresponding pseudo-label $I_k^*$ selected from the memory bank, and $\mathcal{L}_{per}$ is a perceptual loss based on VGG features. 
This formulation ensures that the resulting model is both quality-discriminative and restoration-faithful, balancing perceptual preference with pixel-level accuracy.

\vspace{-1mm}
\section{Experiment}
\vspace{-1mm}
\subsection{Implementation Details}
\vspace{-1mm}

{QualiTeacher is implemented in PyTorch and trained on two NVIDIA A6000 GPUs using AdamW with momentum terms of (0.9, 0.999). The learning rate is set to $5\times10^{-5}$ with only 10K iterations. Since pseudo-labels with scores above 7 are relatively scarce, training a separate score bin for each integer level would lead to insufficient samples and unstable optimization. To address this, we merge all pseudo-labels with quality scores exceeding 7 into a unified high-quality space. Consequently, the maximum guidance score during inference is set to 7.}

During the experiment, we use MUSIQ~\cite{Ke_2021_ICCV}, BRISQUE~\cite{mittal2012no}, and CLIP-IQA~\cite{wang2023exploring} for training.
Since no ground-truth reference images are available for real-world degraded inputs, we employ a comprehensive set of NR-IQA metrics for evaluation across all tasks. These metrics span CNN-based (CNNIQA~\cite{kang2014convolutional}, HYPERIQA~\cite{su2020blindly}, MANIQA~\cite{yang2022maniqa}), Transformer-based (TOPIQ~\cite{chen2024topiq}, ARNIQA~\cite{agnolucci2024arniqa}), and vision-language model-based (Q-Align~\cite{wu2024qalign}, Compare2Score~\cite{zhu2024adaptive}, LIQE~\cite{zhang2023liqe}) methods, where higher values indicate better perceptual quality. In addition, we include task-specific metrics where applicable: FADE~\cite{choi2015referenceless} for dehazing, which measures residual fog density (lower is better); MFRQA~\cite{lin2023noreference} for LLIE, which evaluates brightness, color, structure, and naturalness (higher is better); and URANKER~\cite{guo2023underwater} for UIE (higher is better). The specific subset of metrics used for each task is detailed in \cref{tab:result}.

\vspace{-1mm}
\subsection{Comparative Evaluation}\vspace{-0.5mm}

\begin{figure}[t!]
    \centering
        \setlength{\abovecaptionskip}{0cm} 	\setlength{\belowcaptionskip}{0cm}
    \includegraphics[width=1.0\linewidth]{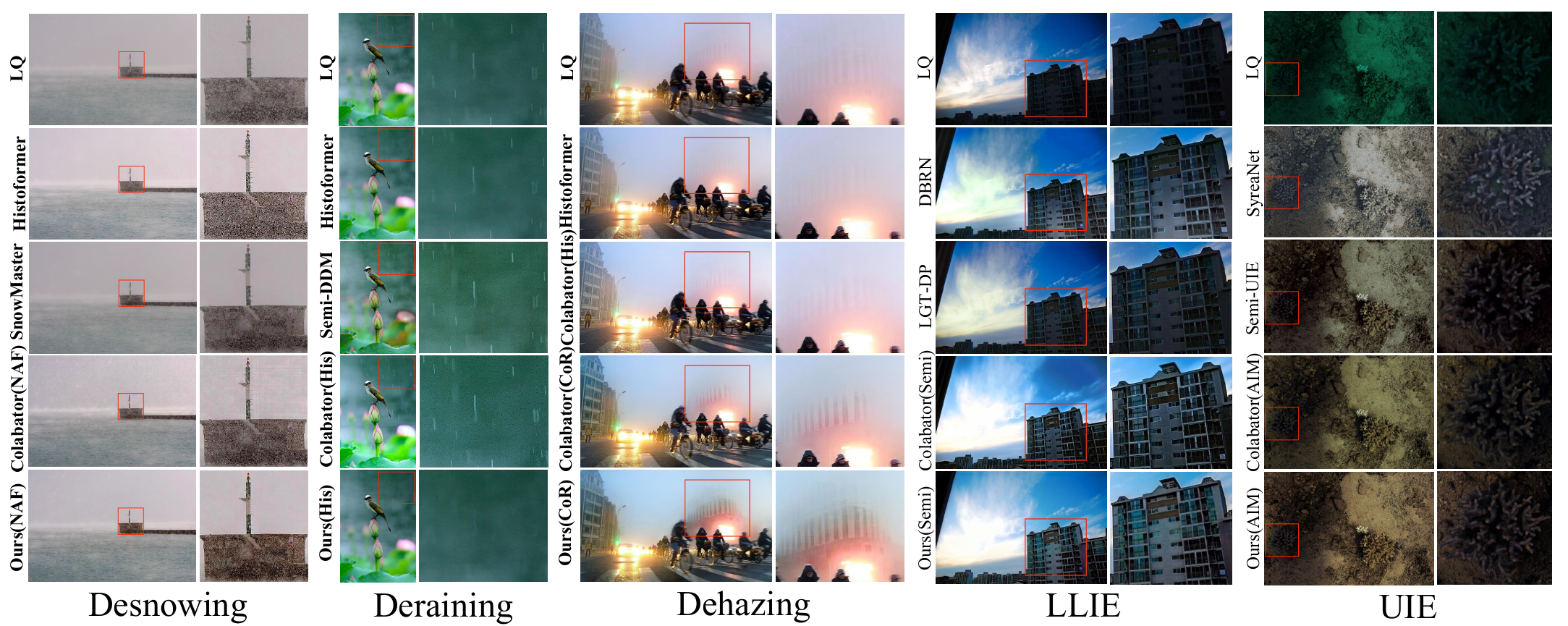}
    \caption{Visualizations on desnowing, deraining, dehazing, LLIE, and UIE.}
    \label{fig:results}
    \vspace{-5mm}
\end{figure}

\begin{table}[t!]
  \centering
  \caption{NR-IQA comparisons on Desnowing, Deraining, Dehazing, LLIE, and UIE.}
  \vspace{-8pt}
  \label{tab:result}
  \resizebox{1.0\textwidth}{!}{
  \setlength{\tabcolsep}{4pt}
  \renewcommand{\arraystretch}{0.92}
  \begin{tabular}{l|cccccccc}
    \toprule
    \midrule
    \rowcolor[RGB]{215,234,236}  \multicolumn{9}{c}{\textbf{Desnowing} }\\
    \midrule
    Methods      & CNNIQA$\uparrow$ & HyperIQA$\uparrow$ & ManIQA$\uparrow$ & Topiq$\uparrow$ & Arniqa$\uparrow$& Q-Align$\uparrow$ & C2score$\uparrow$ & LIQE$\uparrow$ \\
    \midrule
    SemiDDM-Weather~\cite{long2025semiddm} &0.636 & 0.423 &0.342 &0.442 &0.567 &3.586 &55.068 &2.497\\
    \midrule
    SnowMaster (NAFNet)~\cite{lai2025snowmaster} & 0.670       &  0.501        & 0.349       & 0.499      & 0.612       & 3.836        & 55.327        &  2.768    \\
    NAFNet + Colabator~\cite{fang2024real} & 0.628       &  0.470        & 0.342       & 0.473      &  0.620      & 3.936        &  55.389       & 2.667     \\
    NAFNet + QualiTeacher &  \textbf{0.675}  & \textbf{0.559}& \textbf{0.362}& \textbf{0.540}      & \textbf{0.628}       &  \textbf{4.030} & \textbf{55.446}        & \textbf{2.883}\\
    \midrule
    HistoFormer~\cite{sun2024restoring} &0.615 &0.444 &0.346 &0.455 &0.585 &3.827 &55.266 &\textbf{2.610} \\
    HistoFormer + Colabator~\cite{fang2024real} & 0.642  &0.480 &   0.365       & 0.477       &  0.586     & 3.712&55.075 &2.297   \\   
    HistoFormer + QualiTeacher  & \textbf{0.674}       & \textbf{0.530}         & \textbf{0.368}       & \textbf{0.496}      & \textbf{0.592}       & \textbf{ 4.018}       &  \textbf{55.314}       & 2.429     \\
    \midrule
    \midrule
    \rowcolor[RGB]{215,234,236}  \multicolumn{9}{c}{\textbf{Deraining} }\\
    \midrule
    Methods      & CNNIQA$\uparrow$ & HyperIQA$\uparrow$ & ManIQA$\uparrow$ & Topiq$\uparrow$ & Arniqa$\uparrow$ & Q-Align$\uparrow$ & C2score$\uparrow$ & LIQE$\uparrow$ \\
    \midrule
    SSID-KD~\cite{cui2022semi} & 0.580 &0.423 &0.300 &0.427 &0.579 &3.641 &54.790 &2.475\\
    MOSS~\cite{huang2021memory}&0.559 & 0.415& 0.287 &0.416 &0.597 &3.720 &54.803 &2.467\\
    SemiDDM-Weather~\cite{long2025semiddm} &0.620 &0.463 &0.294 &0.438 &0.583 &3.583 &54.712 &2.688\\
    \midrule
    HistoFormer~\cite{sun2024restoring} &0.584 &0.418&0.299 & 0.427 &0.576 &3.734 &54.810 &2.556\\
    HistoFormer + Colabator~\cite{fang2024real} &0.690 &0.545  & 0.308  & \textbf{0.520}  &  0.609&3.690&54.864&\textbf{2.680}  \\   
    HistoFormer + QualiTeacher &\textbf{0.694} &\textbf{0.547} &\textbf{0.308} &0.508 &\textbf{0.613} &\textbf{3.783}&\textbf{54.911}&2.539\\
    \midrule
    \midrule
    \rowcolor[RGB]{215,234,236}  \multicolumn{9}{c}{\textbf{Dehazing}} \\
    \midrule
    Methods      & CNNIQA$\uparrow$ & HyperIQA$\uparrow$ & Topiq$\uparrow$ & Arniqa$\uparrow$ & Q-Align$\uparrow$ & C2score$\uparrow$ & LIQE$\uparrow$ & FADE $\downarrow$\\
    \midrule
    Colabator (CORUN)~\cite{fang2024real} &0.686   &0.602    & 0.599  & 0.718  &3.570   & 55.272   & 2.946  & 0.824   \\
    CORUN + QualiTeacher & \textbf{0.706}  &\textbf{0.621}    &\textbf{0.601}    & \textbf{0.724}   &\textbf{3.628}  &\textbf{55.322}   &\textbf{3.317} &\textbf{0.590}\\
    \midrule
    HistoFormer~\cite{sun2024restoring} & 0.543 &0.415 &0.402 &0.582 &3.050 &54.291 &1.989 &2.499\\
    HistoFormer + Colabator~\cite{fang2024real} &0.690  &\textbf{0.579}   & 0.547  & 0.658  & 3.031&54.363&\textbf{2.473}&1.216  \\   
    HistoFormer + QualiTeacher &\textbf{0.702} &0.550 &\textbf{0.553} &\textbf{0.665} &\textbf{3.270} &\textbf{54.738} &2.460&\textbf{1.145}\\
    \midrule
    \midrule
    \rowcolor[RGB]{215,234,236}  \multicolumn{9}{c}{\textbf{Low-Light Image Enhancement} }\\
    \midrule
    Methods      & CNNIQA$\uparrow$ & ManIQA$\uparrow$ & Topiq$\uparrow$ & Arniqa$\uparrow$ & Q-Align$\uparrow$ & C2score$\uparrow$ & LIQE$\uparrow$ & MFRQA$\uparrow$\\
    \midrule
    DRBN~\cite{yang2020fidelity} &0.564 &0.336 &0.483 &0.652 &3.780 &55.129 &3.136 &56.198\\
    \midrule
    LMT-GP (SemiLL)~\cite{yu2024lmt}  &   0.561 & 0.387  &  0.720 &  0.673 & 3.655 &  55.387 & 3.095 &  58.866    \\
    SemiLL + Colabator~\cite{fang2024real} & 0.614 & 0.372  & 0.724 & 0.671 &  3.730 & 55.103  & 3.320 & 58.138     \\
    SemiLL + QualiTeacher      &\textbf{0.626}   & \textbf{0.411}     &\textbf{0.733}     & \textbf{0.692}    &\textbf{3.804}    & \textbf{55.501}  &\textbf{3.376}   &\textbf{59.372}      \\
    \midrule
    CIDNet~\cite{Yan_2025_CVPR}   &  0.554 & 0.354 &  0.480 &  0.618 & 3.715 &  54.944 & 3.049 &  58.860 \\
    CIDNet + Colabator~\cite{fang2024real}  &0.532  & 0.367 & 0.439       & 0.609      & 3.571       & 54.967        & 2.929        & 59.319     \\
    CIDNet + QualiTeacher  & \textbf{0.669}       & \textbf{0.428}        &\textbf{0.542}        & \textbf{0.657}      &  \textbf{3.741}      & \textbf{55.400}        &  \textbf{3.322}       & \textbf{59.341}     \\
    \midrule
    \midrule
    \rowcolor[RGB]{215,234,236}  \multicolumn{9}{c}{\textbf{Underwater Image Enhancement}} \\
    \midrule
    Methods      & CNNIQA$\uparrow$ & ManIQA$\uparrow$ & Topiq$\uparrow$ & Arniqa$\uparrow$ & Q-Align$\uparrow$ & C2score$\uparrow$ & LIQE$\uparrow$ & URANKER$\uparrow$ \\
    \midrule
    SyreaNet~\cite{wen2023syreanet}&0.449 &0.195 &0.320 &0.513 &3.630 &55.034 &1.658&1.591\\
    \midrule
    Semi-UIR (AIM-Net)~\cite{huang2023contrastive}  &0.438   &0.202   &0.323    &0.474   &3.706   &55.322   &1.642   &2.215 \\
    AIM-Net + Colabator~\cite{fang2024real}  &0.491    &0.244   &0.365   &0.542    &3.825  &55.510  &1.729   &2.374 \\
    AIM-Net + QualiTeacher  &\textbf{0.515}   &\textbf{0.263}   &\textbf{0.376}  &\textbf{0.548}   &\textbf{3.835}   &\textbf{55.639}    &\textbf{1.804}   &\textbf{2.467}  \\   
    \midrule
    \bottomrule
  \end{tabular}
  }
  \vspace{-6mm}
\end{table}

\noindent \textbf{Desnowing.}
With NAFNet as the backbone, we train on paired data from Snow100K~\cite{liu2018desnownet} and unpaired data from RealSnow 10K-Train~\cite{zhu2023Weather}, evaluating on RealSnow 10K-Test~\cite{zhu2023Weather} against SnowMaster~\cite{lai2025snowmaster} and Colabator.
To further probe generalizability, we substitute the backbone with a SOTA model HistoFormer~\cite{sun2024restoring} and benchmark against Colabator~\cite{fang2024real}, alongside the end-to-end semi-supervised method SemiDDM-Weather~\cite{long2025semiddm}.
As reported in \cref{tab:result} and \cref{fig:results}, QualiTeacher consistently outperforms Colabator and existing semi-supervised methods across all metrics. Furthermore, it delivers visually cleaner outputs with sharper textures and fewer residual snow artifacts, confirming the generalization of our quality-prior conditioning strategy to the desnowing task.

\noindent \textbf{Deraining.}
For deraining, we build upon the SOTA weather restoration model HistoFormer~\cite{sun2024restoring} and compare QualiTeacher with Colabator~\cite{fang2024real}, as well as existing semi-supervised baselines SSID-KD~\cite{cui2022semi}, MOSS~\cite{huang2021memory}, and SemiDDM-Weather~\cite{long2025semiddm}.
Both the quantitative results in \cref{tab:result} and the visual comparisons in \cref{fig:results} evaluated on Real 3000~\cite{liu2021unpaired} confirm a clear margin: QualiTeacher significantly outperforms all competing methods on every metric, while better preserving fine-grained textures without introducing over-smoothing artifacts.

\noindent \textbf{Dehazing.}
Following CORUN~\cite{fang2024real}, dehazing performance is assessed on RTTS~\cite{li2018benchmarking}. We additionally swap the backbone to HistoFormer~\cite{sun2024restoring} to test generalizability.
\cref{tab:result} reveals that QualiTeacher achieves the strongest results across all metrics, with particularly pronounced gains on FADE~\cite{choi2015referenceless}, underscoring the capacity of our semi-supervised framework to suppress residual haze.

\noindent \textbf{Low-Light Image Enhancement.}
We adopt SemiLL as the default backbone following LMT-GP~\cite{yu2024lmt}, retaining its original training configuration with VE-LOL~\cite{liu2021benchmarking} for training and DICM~\cite{lee2013contrast} for inference. Comparisons are conducted against Colabator~\cite{fang2024real} and DRBN~\cite{yang2020fidelity}, which do not support backbone replacement. We further substitute the backbone with CIDNet~\cite{Yan_2025_CVPR} to validate generalizability.
As reported in \cref{tab:result}, QualiTeacher leads across all eight metrics. The visualizations in \cref{fig:results} reinforce this advantage: our framework produces more stable and robust enhancements, with notably improved brightness, contrast, and structural fidelity, highlighting the practical value of quality-aware guidance in low-light scenarios.

\noindent \textbf{Underwater Image Enhancement.}
One backbone architecture is employed: AIM-Net from Semi-UIR~\cite{huang2023contrastive}, following the original training configurations of Semi-UIR~\cite{huang2023contrastive}. Inference is performed on Seathru~\cite{akkaynak2019sea}, with Colabator~\cite{fang2024real}, Semi-UIR~\cite{huang2023contrastive} and SyreaNet~\cite{wen2023syreanet}.
\cref{tab:result} and \cref{fig:results} show that QualiTeacher establishes new state-of-the-art results across all metrics, producing underwater images with markedly enhanced color fidelity and finer structural details, demonstrating the effectiveness of quality-aware guidance in this challenging domain.

\begin{table*}[t]
\vspace{-0.5cm}
\begin{minipage}[c]{\textwidth}
  \caption{Break down ablation.}\label{table:breakdown}
  \vspace{-8pt}
\centering
  \resizebox{\linewidth}{!}{
  \setlength{\tabcolsep}{2.3mm}
  \begin{tabular}{c|c|c|c|c|c|c|c|c|c|c}
    \toprule
    ID&   \makecell[c]{Score\\ Condition} & \makecell[c]{Preference \\ Optimization}&  \makecell{ Cropped \\ Consistency} &\makecell[c]{Drop\\ Strategy}  & ManIQA $\uparrow$ & Arniqa $\uparrow$ & HyperIQA $\uparrow$ & Q-Align $\uparrow$ & C2score $\uparrow$  &  LIQE $\uparrow$\\
    \midrule
    (a)&$\times$ &$\times$ &$\times$ &$\times$&0.329 &0.574 & 0.452 &3.850 & 55.309 &2.559 \\
   (b) & $\checkmark$ &$\times$ &$\times$  &$\times$&0.341 &0.601 & 0.510 &4.001 & 55.399 & 2.719\\
   (c) & $\checkmark$ & $\checkmark$ & $\times$ &$\times$&0.352 & 0.613 & 0.535  &4.021 & 55.431 &2.805 \\
   (d)  & $\checkmark$ & $\checkmark$ &$\checkmark$ &$\times$&0.358 & \textbf0.620 & 0.543 & 4.027 & 55.442  & 2.860 \\
      (e)  & $\checkmark$ & $\checkmark$ &$\checkmark$ &$\checkmark$&\textbf{0.362} & \textbf{0.628} & \textbf{0.559} & \textbf{4.030} & \textbf{55.446}&  2.883\\
    \bottomrule
  \end{tabular}}
  \label{tab:ablation}
\end{minipage}\\
\begin{minipage}[c]{\textwidth}
\centering
\setlength{\abovecaptionskip}{0.05cm}
\caption{User study on Desnowing, Dehazing, and LLIE.}\label{table:userstudy} 
\vspace{-10pt}
\resizebox{\textwidth}{!}{
\setlength{\tabcolsep}{2.1mm}
    \begin{tabular}{l cc c l cc c l cc}
        \toprule
        \multicolumn{3}{c}{(a) Desnowing} & &\multicolumn{3}{c}{(b) Dehazing} & &\multicolumn{3}{c}{(c) LLIE} \\
        \cmidrule(lr){1-3} \cmidrule(lr){5-7} \cmidrule(lr){9-11}
        Methods & Score & Rank & & Methods & Score & Rank & & Methods & Score & Rank \\
        \cmidrule(lr){1-3} \cmidrule(lr){5-7} \cmidrule(lr){9-11}
        SemiDDM-Weather~\cite{long2025semiddm} & 2.92 & 4 & &  Colabator (CORUN)~\cite{fang2024real} & 3.35 & 2 & & DRBN~\cite{yang2020fidelity} & 3.67 & 4 \\
        SnowMaster (NAFNet)~\cite{lai2025snowmaster} & 3.42 & 2 & & CORUN + QualiTeacher & \textbf{3.62} & \textbf{1} & & LMT-GP (SemiLL)~\cite{yu2024lmt} & 3.85 & 2 \\
        NAFNet + Colabator~\cite{fang2024real} & 3.23 & 3 & & HistoFormer + Colabator~\cite{fang2024real} & 2.27 & 4 & &   SemiLL + Colabator~\cite{fang2024real} & 3.71 & 3 \\
        NAFNet + QualiTeacher & \textbf{3.58} & \textbf{1} & &  HistoFormer + QualiTeacher & 2.81 & 3 & & SemiLL + QualiTeacher & \textbf{3.96} & \textbf{1} \\
        \cmidrule(lr){1-3} \cmidrule(lr){5-7} \cmidrule(lr){9-11}
        \bottomrule
    \end{tabular}}
\vspace{-5mm}
\end{minipage}
\end{table*}

\vspace{-1mm}
\subsection{Ablation Study}\vspace{-1mm}

We use NAFNet as the backbone, the RealSnow 10K-Test~\cite{zhu2023Weather} as the inference dataset, and conduct ablation studies on the desnowing task. For space limitations, only part ablations are included; please see supplementary material for more results, such as the IQA ensemble strategy and score injection mechanism.

\noindent \textbf{Quality Prior Conditional Approach.}
To steer the student model toward optimal restoration quality, we adopt a quality score of 7 as the conditioning signal. The progression from (a) to (b) in \cref{table:breakdown} confirms that integrating this quality prior into the student network yields consistent improvements. As shown in \cref{fig:ablation}, outputs from (b) present noticeably finer details and snowy artifact removal over (a), validating the benefit of incorporating perceptual priors.

\noindent \textbf{Effectiveness of Preference Optimization.}
Building on the quality prior, preference optimization further elevates the student model's output, as reflected in (b) $\rightarrow$ (c) of \cref{table:breakdown}. As \cref{fig:ablation} shows, quality score conditioning steers the student toward producing higher-quality outputs. However, this IQA-driven optimization can sometimes be a double-edged sword, as it may inadvertently introduce grid-like background artifacts while pursuing higher quality scores.

\noindent \textbf{Effectiveness of Cropped Consistency Loss.}
This component targets the over-optimization problem commonly encountered in continuous score spaces. As shown by (c) $\rightarrow$ (d) in \cref{table:breakdown}, the cropped consistency loss strengthens reconstruction quality while preserving stability and effectively addressing overoptimization. Visual comparisons in \cref{fig:ablation}  echo this observation, revealing refined fine-grained details, enhanced overall perceptual realism, and removed grid-like artifacts in the background.

\noindent \textbf{{Effectiveness of Dual-Drop Mechanism.}}
% The dual-drop mechanism brings a performance improvement, as evidenced by the comparison between (a) and (b) in \cref{table:breakdown}. Visual results in \cref{fig:ablation} further corroborate this finding, where outputs from (b) exhibit sharper edges and richer textures than those of (a). This demonstrates that the dual-drop strategy help the framework's robustness and high-quality. 
The dual-drop mechanism yields a notable performance improvement, as evidenced by the comparison between (d) and (e) in \cref{table:breakdown}. Visual results in \cref{fig:ablation} further corroborate this finding, where outputs from (e) exhibit sharper edges, richer textures, and aesthetic improvement compared to those of (d). This demonstrates that the dual-drop strategy enhances both the robustness and the output quality of the framework.

\subsection{Further Analysis and Extended Applications}
% \noindent \textbf{\textcolor{red}{User Study.}}\quad
% We conduct a user study on the desnowing (RealSnow 10K~\cite{zhu2023Weather}) , dehazing (RTTS~\cite{li2018benchmarking}) tasks and low-light image enhancement tasks (DICM~\cite{lee2013contrast}). Twelve participants rated restored images on a 0-5 scale (0 = worst, 5 = best) based on three criteria: visibility, artifact removal, and naturalness, assessing the extent to which weather-related degradations are eliminated while preserving realistic appearance. For fairness, each degraded image and its restored counterpart were displayed side by side, with method identities hidden. As shown in \cref{table:userstudy}, our method consistently receives the highest ratings, confirming its superiority in perceptual quality.

% \noindent \textbf{Benefits for Downstream Tasks.}\quad
% Improving image quality can facilitate various downstream vision tasks. To validate this, we evaluate how different enhancement methods affect low-light object detection. Specifically, we enhance the low-light images from ExDark~\cite{loh2019getting} using various algorithms and feed the enhanced results into YOLO for detection. We also include a baseline that directly performs detection on the original low-light inputs without any enhancement. As reported in \cref{tab:exdark}, QualiTeacher achieves the best detection accuracy among all methods, and consistently boosts performance when integrated with other restoration frameworks.

\noindent \textbf{User Study.}
We conduct a user study on desnowing (RealSnow 10K~\cite{zhu2023Weather}), dehazing (RTTS~\cite{li2018benchmarking}), and LLIE (DICM~\cite{lee2013contrast}). Fifteen participants rated restored images on a 0-5 scale based on visibility, artifact removal, and naturalness. Each degraded image and its restored counterpart were displayed side by side, with method identities hidden. As shown in \cref{table:userstudy}, our method consistently receives the highest ratings, confirming its superiority in perceptual quality.
\begin{figure*}[t!]
   \begin{minipage}[c]{0.56\textwidth}
\centering
  \setlength{\abovecaptionskip}{-0.0cm}
\includegraphics[width=\textwidth]{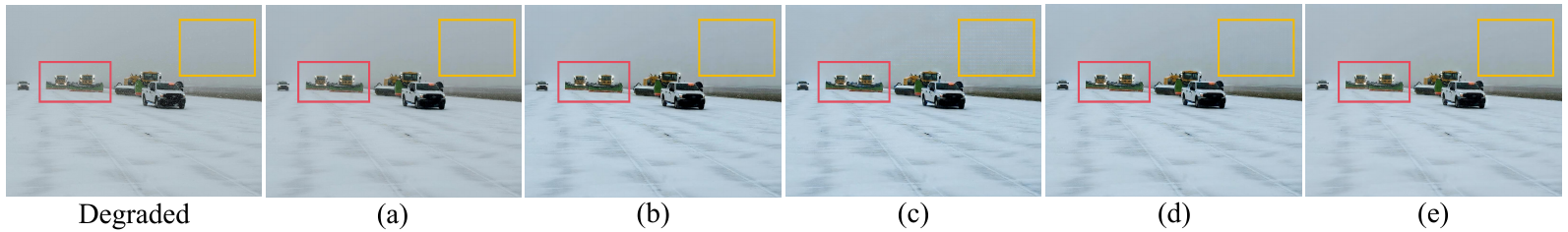}
% \vspace{-5pt}
	\caption{Visualization of breakdown ablation, where (a), (b), (c), (d), and (e) are consistent with those in \cref{tab:ablation}.}
	\label{fig:ablation}
	% \vspace{-5.5mm}
\end{minipage}
\begin{minipage}[c]{0.42\textwidth}
\centering
  \setlength{\abovecaptionskip}{-0.0cm}
\includegraphics[width=\textwidth]{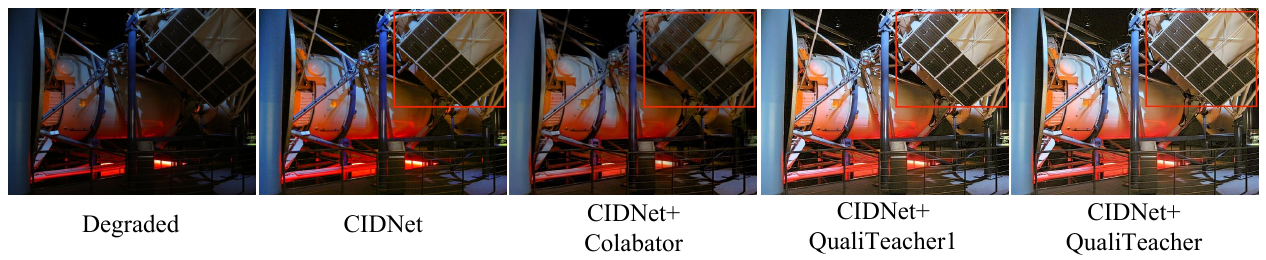} 
% \vspace{-3mm}
	\caption{Visualization on LLIE with changed IQA selection (MANIQA, TOPIQ, QualiClip).}
	\label{fig:threeiqa}
    \end{minipage}
	\vspace{-4mm}
\end{figure*}
\begin{table}[t!]
%\vspace{-}
  \centering
  \caption{Ablation on the IQA selection. QualiTeacher with alternative NR-IQA metrics (MANIQA, TOPIQ, QualiClip) on LLIE.}\label{tab.IQAselection}
  \vspace{-10pt}
  \label{tab:nriqa}
  \resizebox{\textwidth}{!}{%
  \begin{tabular}{l|cccccccc}
    \toprule
    % \rowcolor[RGB]{215,234,236}  \multicolumn{9}{c}{\textbf{Desnowing} }\\
    % \midrule
    Methods      & CNNIQA$\uparrow$ & ManIQA$\uparrow$ & Topiq$\uparrow$ & Arniqa$\uparrow$& Q-Align$\uparrow$ & C2score$\uparrow$ & LIQE$\uparrow$ & MFRQA$\uparrow$ \\
    \midrule
    %  DRBN~\cite{yang2020fidelity} &0.564 &0.336 &0.483 &0.652 &3.780 &55.129 &3.136 &56.198\\
    % \midrule
    %  LMT-GP (SemiLL)~\cite{yu2024lmt}  &   0.561 & 0.387  &  0.720 &  0.673 & 3.655 &  55.387 & 3.095 &  58.866    \\
     CIDNet~\cite{Yan_2025_CVPR}   &  0.554 & 0.354 &  0.480 &  0.618 & 3.715 &  54.944 & 3.049 &  58.860 \\
    CIDNet + Colabator~\cite{fang2024real}  &0.532  & 0.367 & 0.439       & 0.609      & 3.571       & 54.967        & 2.929        & 59.319     \\
    CIDNet + QualiTeacher1 &\secon{0.660} &\first{0.432} & \secon{0.530} &\first
{0.661} &\secon{3.643} &\first{55.454} &\first{3.411} & \first{59.675} \\
CIDNet + QualiTeacher(Ours)  & \first{0.669}       & \secon{0.428}        &\first{0.542}        & \secon{0.657}      &  \first{3.741}      & \secon{55.400}        &  \secon{3.322}       &\secon{59.341}     \\
   %  \midrule
   %  \rowcolor[RGB]{215,234,236}  \multicolumn{9}{c}{\textbf{Underwater Image Enhancement} }\\
   %  \midrule
   %   Methods      & CNNIQA$\uparrow$ & HyperIQA$\uparrow$ & ManIQA$\uparrow$ & Topiq$\uparrow$ & Arniqa$\uparrow$& Q-Align$\uparrow$ & C2score$\uparrow$ & LIQE$\uparrow$ \\
   %  \midrule
   % Semi-UIR (AIM-Net)~\cite{huang2023contrastive}  &   &   &    &   &   &   &   & \\
   %  AIM-Net + Colabator~\cite{fang2024real}  &    &   &   &    &  &  &   & \\
   %  AIM-Net + QualiTeacher  &   &   &  &   &   &     &   &  \\   
    \bottomrule
  \end{tabular}
  }
  \vspace{-0.6cm}
\end{table}

\noindent \textbf{{IQA Selection.}}
Our IQA selection follows a complementary-level principle, combining low-level and high-level metrics for comprehensive quality supervision. To verify that performance gains stem from our quality-prior mechanism rather than specific IQA choices, we substitute three alternative NR-IQA models (MANIQA~\cite{yang2022maniqa}, TOPIQ~\cite{chen2024topiq}, and QualiClip~\cite{agnolucci2024quality}) under the same principle and retrain the framework to get QualiTeacher1. As shown in~\cref{tab.IQAselection} and ~\cref{fig:threeiqa}, evaluation on LLIE with eight metrics under identical settings still achieves state-of-the-art performance, confirming that improvement is attributed to our strategy design rather than particular IQA engineering selection as well as verifying the generalizability of our quality-prior mechanism.

\begin{figure*}[t!]
\vspace{-3pt}
\begin{minipage}[c]{0.51\textwidth}
\centering
  \setlength{\abovecaptionskip}{-0.0cm}
\includegraphics[width=\textwidth]{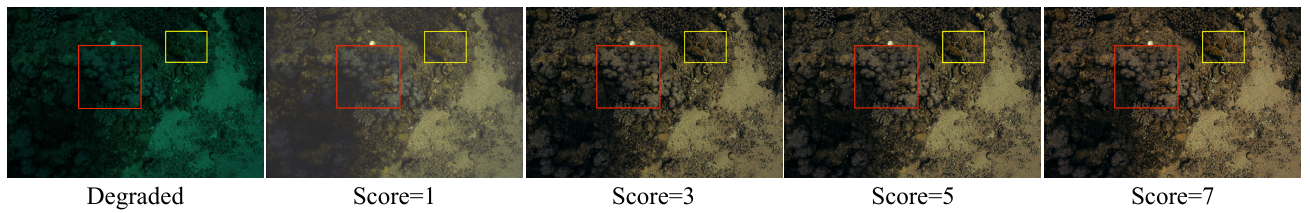} 
% \vspace{-3mm}
	\caption{Score controllability.}
	\label{fig:score controllability}
    \end{minipage}
    \begin{minipage}[c]{0.47\textwidth}
\centering
  \setlength{\abovecaptionskip}{-0.0cm}
\includegraphics[width=\textwidth]{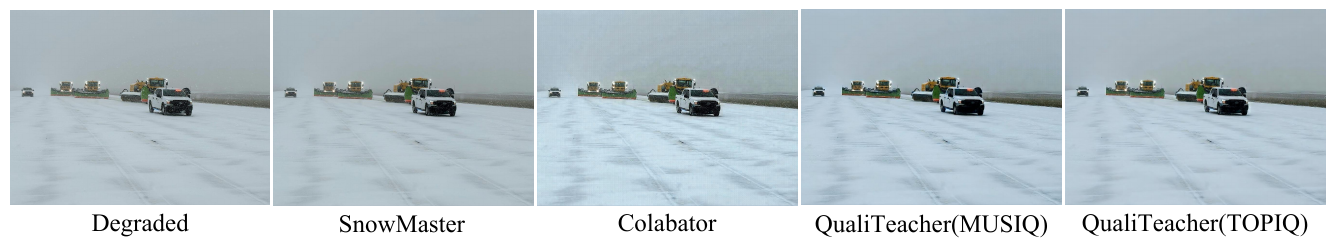}
% \vspace{-5pt}
	\caption{Failure cases analysis.}
	\label{fig:limitation}
	% \vspace{-5.5mm}
\end{minipage}\\
	\vspace{-6mm}
\end{figure*}

 \noindent \textbf{Score Controllability Analysis.}
We show that the condition score directly influences restoration quality at test time. As illustrated in \cref{fig:score controllability}, by varying the injected score at inference, the model generates outputs of corresponding quality, enabling continuous control over perceptual quality. In practice, we set the condition to the maximum score to achieve optimal restoration results. Notably, quality controllability proves particularly beneficial for UIE, where higher score guidance yields more visually balanced and realistic color tones.
 
\noindent \textbf{Benefits for Downstream Tasks.}
To validate that improving image quality facilitates downstream tasks, we evaluate enhancement methods on low-light object detection. We enhance low-light images from ExDark~\cite{loh2019getting} using various algorithms and feed the results into YOLOv13~\cite{lei2025yolov13} for detection. As reported in \cref{tab:exdark}, QualiTeacher consistently boosts detection performance.
% when integrated with restoration methods.

\begin{table*}[t]
\centering
\caption{{Low-light image detection on }\textit{ExDark}.}\label{tab:exdark}
\vspace{-10pt}
\setlength{\tabcolsep}{4pt}
\resizebox{\textwidth}{!}{
\begin{tabular}{l|cccccccccccc|c}
    \toprule
    Methods (AP) & Bicycle & Boat & Bottle & Bus & Car & Cat & Chair & Cup & Dog & Motor & People & Table & Mean \\
    \midrule
    DRBN~\cite{yang2020fidelity} &81.8                         & 81.1                         & 73.5                         & 90.2                         & 79.6                         & 72.6                         & 65.1                         & 69.9                         & 78.2                         & 71.2                         & 76.7                         & 62.7                         & 75.2 \\
    LMT-GP(SemiLL)~\cite{yu2024lmt} & 80.9                         & 82.7                         & 76.2                         & \secon{91.2}                         & 80.2                         & 75.1                         & 72.1                         & 70.6                         & 77.8                         & 71.1                         & 78.8                         & 65.8                         & 76.9 \\
    SemiLL + Colabator~\cite{fang2024real} &79.6                         & 82.0                         & 75.7                         & 89.2                         & 80.8                         & 69.5                         & 70.0                         & 70.8                         & 79.9                         & 71.6                         & 76.8                         & 65.6                         & 76.0 \\
    CIDNet + Colabator~\cite{fang2024real}  &80.7                         & 79.8                         & 76.0                         & 86.6                         & 81.2                         & \secon{75.2}                         & 71.9                         & 71.9                         & 79.6                         & 72.1                         & 78.3                         & 67.3                         & 76.7 \\
    \rowcolor{gray!10}  SemiLL + QualiTeacher &\secon{82.6} &\first{83.3} &\first{78.2} &\secon{91.2} &\secon{82.1} &70.9 &\first{74.6} &\secon{72.0}&\secon{80.7} &\first{74.2} &\secon{80.5} & \secon{67.9} &\secon{78.2}  \\
    \rowcolor{gray!10} CIDNet + QualiTeacher & \first{82.9} &\secon{83.0} &\secon{76.4} &\first{92.5} &\first{82.5} & \first{77.8} & \secon{73.3} &\first{73.5} &\first{82.8}&\secon{74.0} &\first{81.4} &\first{68.7}& \first{79.1}  \\
    \bottomrule
\end{tabular}}
\vspace{-6pt}
\end{table*}
\begin{table}[t!]
\vspace{-5pt}
  \centering
  \caption{Robust analysis for NR-IQA metrics. We replace MUSIQ with TOPIQ.}\label{Fig.musiq}
  \vspace{-10pt}
  \label{tab:limit}
  \resizebox{\textwidth}{!}{%
  \setlength{\tabcolsep}{1.5mm}
  \begin{tabular}{l|cccccccc}
    \toprule
    % \rowcolor[RGB]{215,234,236}  \multicolumn{9}{c}{\textbf{Desnowing} }\\
    % \midrule
    Methods (Task: Desnowing)      & CNNIQA$\uparrow$ & HyperIQA$\uparrow$ & ManIQA$\uparrow$ & Topiq$\uparrow$ & Arniqa$\uparrow$& Q-Align$\uparrow$ & C2score$\uparrow$ & LIQE$\uparrow$ \\
    \midrule
    SnowMaster(NAFNet)~\cite{lai2025snowmaster} & 0.670       &  0.501        & 0.349       & 0.499      & 0.612       & 3.836        & 55.327        &  2.768    \\
    NAFNet + Colabator~\cite{fang2024real} & 0.628       &  0.470       & 0.342       & 0.473      &  0.620      & 3.936        &  55.389       & 2.667     \\
    \rowcolor{gray!10} NAFNet + QualiTeacher(Topiq) &  \secon{0.671}  & \secon{0.538}& \secon{0.361}& \first{0.540}      & \secon{0.614}       & \secon{3.999} & \secon{55.398} & \secon{2.826}\\
   \rowcolor{gray!10} NAFNet + QualiTeacher &  \first{0.675}  & \first{0.559}& \first{0.362}& \first{0.540}      & \first{0.628}       &  \first{4.030} & \first{55.446}  
    & \first{2.883}\\
    \bottomrule
  \end{tabular}
  }
  \vspace{-3mm}
\end{table}

\vspace{-2mm}
\section{Limitations}\vspace{-2mm}
As shown in \cref{fig:limitation}, QualiTeacher occasionally produces grid-like artifacts in a very small number of images, while Colabator exhibits them in a considerably larger portion of its results. Since reducing training iterations only mitigates but does not eliminate these artifacts, we hypothesize that the root cause lies in the MUSIQ-KonIQ metric employed by both methods. Although our framework adopts an ensemble strategy to minimize single-metric bias, MUSIQ is trained on KonIQ~\cite{koniq10k}, a photography-oriented dataset whose high-quality references tend to contain sharp, repetitive texture patterns. Consequently, MUSIQ may assign higher scores to outputs with grid-like structures, inadvertently guiding the model toward such artifacts. To verify this, we replace Musiq with Topiq~\cite{chen2024topiq}. As shown in \cref{Fig.musiq} and \cref{fig:limitation}, the grid artifacts vanish, confirming our conjecture. Notably, QualiTeacher (TOPIQ) still achieves SOTA performance, demonstrating our framework's robustness to the choice of IQA metrics.

\vspace{-1mm}
\section{Conclusion}\vspace{-1mm}
In this paper, we propose QualiTeacher, a novel framework that reframes pseudo-label quality as a conditional signal instead of a filter. This quality-conditioned approach enables our student model to learn a complete restoration manifold, which critically allows it to extrapolate beyond the teacher's quality ceiling and achieve state-of-the-art results. Our work, proven robust across various NR-IQA models, establishes a new paradigm for learning from imperfect supervision and opens new avenues for flexible restoration strategies.

\bibliographystyle{splncs04}
\bibliography{main}
\end{document}